\documentclass{article}
\usepackage{spconf,amsmath,graphicx,hyperref}
\usepackage{amsthm}      % 定义 proof 环境、\newtheorem 等
\usepackage{cleveref}    % 给 \cref、\Cref 等智能交叉引用命令
\usepackage{caption}     % 支持 \captionsetup{…}
\usepackage{mdframed}    % 支持 mdframed 环境
\definecolor{headergray}{gray}{0.85}
\definecolor{rowgray}{gray}{0.95}
\definecolor{bestgreen}{rgb}{0.0, 0.6, 0.0}

\usepackage{tabularx}
\usepackage{graphicx}
\usepackage{textcomp}
\usepackage{xcolor}
\usepackage{multirow}
\usepackage{listings}
\usepackage[utf8]{inputenc}
\usepackage[T1]{fontenc}
\usepackage{hyperref}
\usepackage{url}
\usepackage{booktabs}
\usepackage{nicefrac}
\usepackage{microtype}
\usepackage{algorithm}
\usepackage{algpseudocode}
\usepackage{adjustbox}
\usepackage{tcolorbox}
\usepackage{amsthm}      % 定义 proof 环境、\newtheorem 等
\usepackage{cleveref}    % 给 \cref、\Cref 等智能交叉引用命令
\usepackage{caption}     % 支持 \captionsetup{…}
\usepackage{mdframed}    % 支持 mdframed 环境
\title{SCORE: Story Coherence and Retrieval Enhancement for AI Narratives}
\name{%
\begin{tabular}{@{}c@{}} 
Qiang Yi$^{1*}$, Yangfan He$^{2*}$, Jianhui Wang$^{3*}$ Xinyuan Song$^{4}$, ShiYao Qian$^{5}$, Xinhang Yuan$^{6}$,
Yi Xin$^{7}$, \\ Yijin Wang$^{8}$, Jingqun Tang$^{9}$, Yuchen Li$^{10}$, Junjiang Lin$^{17}$ Hongyang He$^{11}$, Zhen Tian$^{12}$, Tianxiang Xu$^{19}$, \\ 
Keqin Li$^{13}$, Kuan Lu$^{14}$, Menghao Huo$^{15}$, Jiaqi Chen$^{14}$, Miao Zhang$^{16}$, Tianyu Shi$^{5}$, Jianyuan Ni$^{18\dagger}$ 
\end{tabular} 
\thanks{* Equal contribution. † Corresponding author.} 
} 
\address{$^{1}$ UCB, $^{2}$ UMN, $^{3}$ UESTC, $^{4}$ Emory, $^{5}$ UofT, $^{6}$ WUSTL, $^{7}$ NJU, $^{8}$ XDU,
$^{9}$ ByteDance, $^{10}$ Baidu, \\  $^{11}$ UofWarwick, $^{12}$ UofGlasgow, $^{13}$ AMA, $^{14}$ Google, $^{15}$ SCU, $^{16}$ THU-SZ, $^{17}$ Amazon, $^{18}$ JC, $^{19}$ PKU}

\begin{document}
%\ninept
%
\maketitle
\begin{abstract}
Large Language Models (LLMs) can generate creative and engaging narratives from user-specified input, but maintaining coherence and emotional depth throughout these AI-generated stories remains a challenge. In this work, we propose SCORE, a framework for Story Coherence and Retrieval Enhancement, designed to detect and resolve narrative inconsistencies. By tracking key item statuses and generating episode summaries, SCORE uses a Retrieval-Augmented Generation (RAG) approach to identify related episodes and enhance the overall story structure. Experimental results from testing multiple LLM-generated stories demonstrate that SCORE significantly improves the consistency and stability of narrative coherence compared to baseline GPT models, providing a more robust method for evaluating and refining AI-generated narratives.
\end{abstract}


\section{Introduction}
Deep learning has transformed multiple domains including NLP, time series analysis and computer vision~\cite{qiu2025easytime, qiu2025duet, qiu2024tfb, li2024towards, li2023bilateral,  li2024distinct}. Large Language Models (LLMs) have demonstrated significant capabilities in generating long-form narratives, such as serialized stories or novels, by leveraging large-scale architectures and vast amounts of training data ~\cite{tao2024rolecraft,brown2020language, touvron2023llama}. However, maintaining narrative consistency over extended texts, especially in terms of character development and emotional coherence, remains a major challenge~\cite{mcadams2006problem}. For instance, \cite{khatun2024assessing} pointed out that achieving thematic consistency and managing dynamic plot states is crucial for maintaining the logical flow of a story. In practice, LLMs often struggle with inconsistencies when characters or key plot items reappear without proper explanation, disrupting the overall narrative structure~\cite{celikyilmaz2020evaluation}.

Similarly, \cite{liu2024generating} highlight the difficulties in managing multimodal elements within long-form narratives, noting that inconsistencies in character behavior or emotional tone can negatively impact reader engagement. These challenges indicate a need for more structured approaches in narrative generation that can better manage character arcs, plot developments, and emotional progression throughout the story.

\begin{figure*}[t]
\centering
\includegraphics[width=1\linewidth]{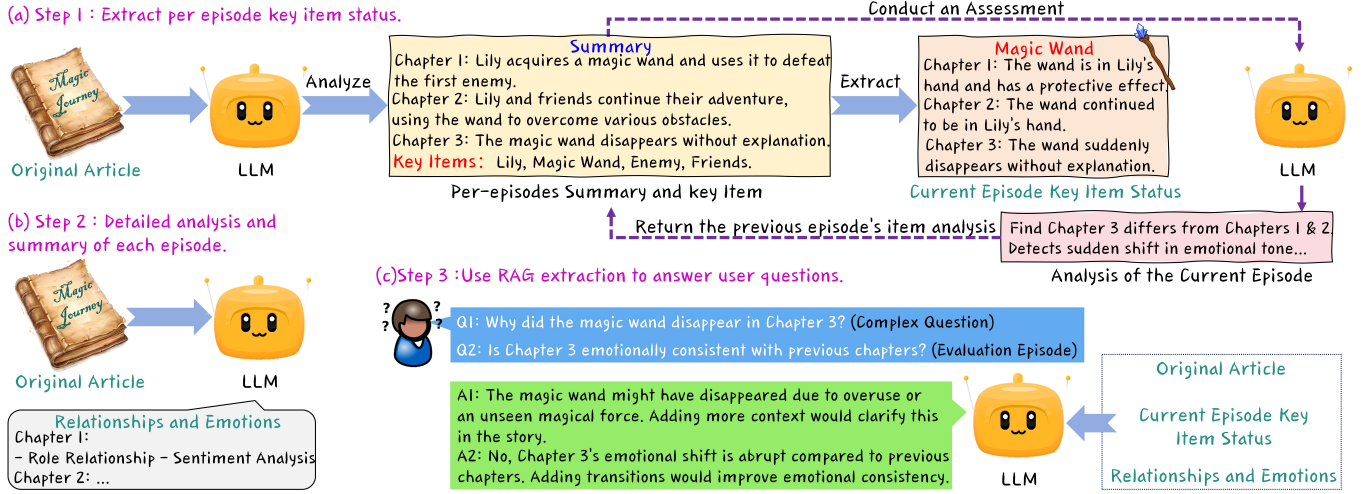}
\caption{\textbf{Overview of SCORE}: (a) Extracts key item statuses per episode. (b) Conducts detailed analysis and summaries of each episode. (c) Uses RAG to answer user queries and resolve narrative inconsistencies.}
\label{fig:method}
\end{figure*}

In addition, recent works have highlighted the importance of memory mechanisms in LLM-based agents. \cite{zhang2024survey} conducted a comprehensive survey on these mechanisms, identifying effective memory designs that help mitigate inconsistencies in narrative development—a challenge common to both interactive agents and narrative generation tasks. Additionally, \cite{park2023generativeagentsinteractivesimulacra} introduced generative agents that simulate human-like behavior using memory modules. These agents track the state of a wide array of interactable objects in a sandbox environment, ensuring consistent reasoning and enabling the smooth functioning of a simulated society. These researches inspired the design of our new framework.

In this work, we build upon recent advancements in Retrieval-Augmented Generation (RAG)~\cite{lewis2020retrieval}, which dynamically incorporates relevant context to enhance narrative coherence. Expanding on these developments, we propose SCORE, a framework designed to evaluate three critical aspects of long-form narrative generation: character consistency, emotional coherence, and logical tracking of key plot elements. Our key contributions are:

\begin{itemize} \item We introduce SCORE, an LLM-based evaluation framework that detects narrative inconsistencies in AI-generated stories. \item We incorporate a Retrieval-Augmented Generation (RAG) approach, utilizing episode-level summaries and key item tracking to improve narrative coherence. \item We demonstrate enhanced story consistency and emotional depth by integrating sentiment analysis and similarity-based episode retrieval. Specially, we outperform baseline GPT model~\cite{sradford2019language} in detecting continuity errors and maintaining overall narrative coherence. 
\end{itemize}

\section{Method}

Our proposed method, SCORE, consists of three main components: (1) an LLM-based evaluation framework to assess the coherence of key story elements, (2) automatic generation of episode summaries to track plot development, and (3) a retrieval-augmented generation (RAG) approach that integrates the first two components, enabling enhanced user interaction and ensuring narrative consistency. 

As the framework is intended solely for academic research purposes, its use is consistent with the original access conditions of all incorporated tools and data sources.

\subsection{Continuity Analysis and Key Item Status Correction}
By extracting key parts of GPT-4's analysis, we identify instances where an item reappears in later episodes after being marked as lost or destroyed, without narrative explanation. Let $S_i(t)$ represent the state of item $i$ at time $t$, where $S_i(t) \in \{\text{active}, \text{lost}, \text{destroyed}\}$. If item $i$ reappears at time $t_k$ with $S_i(t_k) = \text{active}$ after being previously marked as $S_i(t_{k-1}) \in \{\text{lost}, \text{destroyed}\}$, we flag this as a continuity error. To maintain consistency, the state remains $S_i(t_{k-1})$, avoiding an incorrect update. This approach systematically corrects discrepancies in item states, ensuring that narrative continuity is preserved by preventing erroneous state transitions.

\subsection{Key Item Interaction Analysis}

For each episode, we conduct a thorough evaluation by summarizing key plot points, character actions, and tracking interactions with important items. Let $A_c(t)$ represent the actions of character $c$ at time $t$, and let $I_i(t)$ denote interactions with key item $i$. The model generates summaries that encapsulate essential elements, including $A_c(t)$ (character actions), relationships, and emotional changes across the episode. It then analyzes the specific interactions $I_i(t)$ between characters and key items, documenting these for further analysis. This step aggregates relevant narrative information—combining episode summaries, key item interactions $I_i(t)$, and character actions $A_c(t)$—to facilitate more precise future retrieval. The approach simplifies subsequent analysis of plot and item continuity, reducing redundancy and improving efficiency.

\begin{table*}[t!]
\centering
\caption{Performance Comparison of LLMs with/without SCORE Framework}
\label{tab:performance}
\begin{tabularx}{\textwidth}{l*{4}{>{\centering\arraybackslash}X}}
\toprule
\textbf{Model} & \textbf{Consistency} & \textbf{Coherence} & \textbf{Item Status} & \textbf{Complex QA} \\
\midrule
GPT-4 & 83.21 & 84.32 & 0 & 82.34 \\
SCORE & \textbf{85.61} (\textcolor{bestgreen}{$\uparrow$2.4}) & \textbf{86.9} (\textcolor{bestgreen}{$\uparrow$2.58}) & \textbf{98} (\textcolor{bestgreen}{$\uparrow$98}) & \textbf{89.45} (\textcolor{bestgreen}{$\uparrow$7.11}) \\
\midrule
GPT-4o & 86.78 & 82.21 & 0 & 76.32 \\
SCORE & \textbf{88.68} (\textcolor{bestgreen}{$\uparrow$1.9}) & \textbf{89.91} (\textcolor{bestgreen}{$\uparrow$7.7}) & \textbf{96} (\textcolor{bestgreen}{$\uparrow$96}) & \textbf{88.75} (\textcolor{bestgreen}{$\uparrow$12.43}) \\
\midrule
Claude3 & 84.6 & 80.9 & 0 & 69.45 \\
SCORE & \textbf{87.2} (\textcolor{bestgreen}{$\uparrow$2.6}) & \textbf{85.7} (\textcolor{bestgreen}{$\uparrow$4.8}) & \textbf{93.1} (\textcolor{bestgreen}{$\uparrow$93.1}) & \textbf{83.9} (\textcolor{bestgreen}{$\uparrow$14.45}) \\
\midrule
Gemini-Pro & 82.2 & 83.4 & 0 & 78.40 \\
SCORE & \textbf{85.2} (\textcolor{bestgreen}{$\uparrow$3.2}) & \textbf{86.0} (\textcolor{bestgreen}{$\uparrow$3.0}) & \textbf{95.0} (\textcolor{bestgreen}{$\uparrow$95.0}) & \textbf{85.5} (\textcolor{bestgreen}{$\uparrow$7.10}) \\
\midrule
Qwen-14B & 79.3 & 76.2 & 0 & 25.15 \\
SCORE & \textbf{84.5} (\textcolor{bestgreen}{$\uparrow$5.2}) & \textbf{79.1} (\textcolor{bestgreen}{$\uparrow$2.9}) & \textbf{86.3} (\textcolor{bestgreen}{$\uparrow$86.3}) & \textbf{61.2} (\textcolor{bestgreen}{$\uparrow$36.05}) \\
\midrule
GPT-4o-mini & 77.5 & 75.8 & 0 & 23.40 \\
SCORE & \textbf{81.9} (\textcolor{bestgreen}{$\uparrow$4.4}) & \textbf{76.6} (\textcolor{bestgreen}{$\uparrow$0.8}) & \textbf{79.8} (\textcolor{bestgreen}{$\uparrow$79.8}) & \textbf{60.1} (\textcolor{bestgreen}{$\uparrow$36.7}) \\
\midrule
Llama-13B & 71.3 & 69.8 & 0 & 18.72 \\
SCORE & \textbf{79.1} (\textcolor{bestgreen}{$\uparrow$7.8}) & \textbf{73.4} (\textcolor{bestgreen}{$\uparrow$3.6}) & \textbf{76.2} (\textcolor{bestgreen}{$\uparrow$76.2}) & \textbf{57.3} (\textcolor{bestgreen}{$\uparrow$38.58}) \\
\bottomrule
\end{tabularx}
\vspace{-15pt}
\end{table*}

\subsection{Similarity-Based Episode Evaluation and Sentiment Analysis}
We integrate similarity-based retrieval and sentiment analysis to improve episode evaluation and answer complex queries. It begins by loading summaries, full episode content, and key item states from structured JSON files. The content is segmented into smaller chunks using a text segmenter and embedded into a vector space model using FAISS~\cite{douze2024faisslibrary} and OpenAI embeddings. This vector space enables efficient retrieval of similar episodes for user queries or specific episode analysis.

For evaluation, the system retrieves relevant past episodes by calculating cosine similarity scores~\cite{rahutomo2012semantic} between the current episode or query and all other episodes in the vector space. Let $S(e_c, e_p)$ represent the similarity score between the current episode $e_c$ and a past episode $e_p$. The top $N$ episodes with the highest $S(e_c, e_p)$ scores are retrieved for further analysis, providing a relevant summary of episodes for evaluation or answering questions.

Sentiment analysis is then applied to both the current and retrieved episodes. A sentiment score $\sigma(e)$, ranging from 0 to 1, is assigned to each episode $e$ by GPT-4, reflecting its emotional tone. These scores help refine the selection by ensuring both text similarity and sentiment consistency are considered, thus preventing errors from large sentiment discrepancies.

Finally, the LLM processes the retrieved episode summaries and content to generate a detailed evaluation. The focus is on narrative aspects such as character consistency, plot progression, emotional authenticity, and key item continuity. This ensures the narrative remains coherent, with any discrepancies flagged and corrected.

\section{Experiments}
To evaluate the effectiveness of SCORE, we conducted experiments on stories generated by large language models (LLMs). These experiments assessed the framework's ability to maintain narrative coherence, detect continuity errors, and ensure emotional consistency throughout episodic storytelling.

\begin{table*}[t]
\centering
\caption{Ablation results for our complete configuration (Full SCORE) versus removing each module.}
\label{tab:ablation}
\begin{tabular}{lcccc}
\toprule
\textbf{Configuration} & \textbf{Consistency} & \textbf{Coherence} & \textbf{Item Acc.} & \textbf{Complex QA} \\
\midrule
Full SCORE & 89.7 & 91.2 & 98.3 & 88.5 \\
w/o Dynamic Tracking & 72.1 (\textcolor{bestgreen}{$\downarrow$17.6}) & 85.4 (\textcolor{bestgreen}{$\downarrow$5.8}) & 61.2 (\textcolor{bestgreen}{$\downarrow$37.1}) & 67.3 (\textcolor{bestgreen}{$\downarrow$21.2}) \\
w/o Context Summary & 83.2 (\textcolor{bestgreen}{$\downarrow$6.5}) & 68.7 (\textcolor{bestgreen}{$\downarrow$22.5}) & 89.1 (\textcolor{bestgreen}{$\downarrow$9.2}) & 71.4 (\textcolor{bestgreen}{$\downarrow$17.1}) \\
w/o Hybrid Retrieval & 86.4 (\textcolor{bestgreen}{$\downarrow$3.3}) & 88.9 (\textcolor{bestgreen}{$\downarrow$2.3}) & 94.2 (\textcolor{bestgreen}{$\downarrow$4.1}) & 77.6 (\textcolor{bestgreen}{$\downarrow$10.9}) \\
w/o Sentiment & 87.1 (\textcolor{bestgreen}{$\downarrow$2.6}) & 89.5 (\textcolor{bestgreen}{$\downarrow$1.7}) & 96.8 (\textcolor{bestgreen}{$\downarrow$1.5}) & 82.3 (\textcolor{bestgreen}{$\downarrow$6.2}) \\
\bottomrule
\end{tabular}
\vspace{-10pt}
\end{table*}

\begin{figure*}[!t]
\centering
\includegraphics[width=0.8\textwidth]{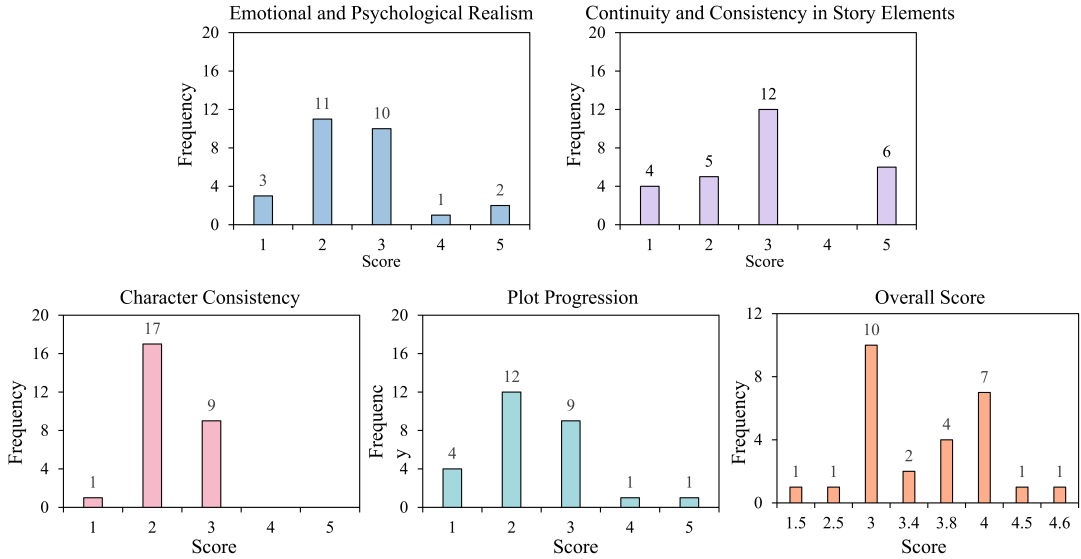}
\caption{Case study.}
\label{fig:scores_distribution}
\vspace{-15pt}
\end{figure*}

\subsection{Dataset Preparation}
We constructed a story dataset of 5,000 episodes from 1,000 GPT-generated stories evenly distributed across four genres (science fiction, drama, fantasy, comedy). Each story contains 10–15 episodes (avg. 12, $\sim$2,000 tokens), generated using GPT-3.5 and GPT-4 outputs with 15 diverse prompt templates per genre. Quality was ensured via manual filtering by three annotators ($\kappa$=0.78), and complexity stratification included multi-character interactions (30\%), non-linear timelines (20\%), and symbolic systems (50\%). For broader validation, SCORE was also tested on NarrativeQA, BookCorpus, and WP-STORIES, with genre and episode length distributions standardized for cross-dataset comparison. This hybrid setup balances synthetic and human-authored narratives, ensuring robustness across industrial and academic benchmarks.

\subsection{Metrics}

We evaluated the framework using several key metrics: \textbf{narrative coherence}, assessed by examining logical consistency in character behavior and plot development to avoid continuity errors; \textbf{consistency}, defined as the fraction of responses not conflicting with prior information; \textbf{coherence}, measured as the average semantic or logical flow score across responses; \textbf{item status}, calculated as the proportion of required elements present in each response; and \textbf{complex question}, defined as the fraction of correctly answered complex questions. Finally, we compared the evaluation scores from different methods to measure the stability of the framework.

\subsection{Evaluation Process}

For each episode, evaluation proceeded in two stages. First, we directly uploaded files to ChatGPT (GPT-4o-mini, GPT-4o, GPT-4) to test baseline narrative assessment without prompts. Second, we performed detailed evaluation using preprocessed files with RAG support, where FAISS-based cosine similarity retrieved semantically and emotionally aligned episodes to provide context. This enriched context was then incorporated into GPT prompts for more accurate key-item tracking and narrative evaluation.

\subsection{Baselines}

We compared our proposed framework to three baselines: GPT-4, GPT-4o, and GPT-4o-mini. In these cases, we used these models directly without integrating our key item tracking, continuity analysis, or retrieval-augmented generation (RAG) mechanisms. We used the same LLM-generated stories to evaluate different models. For all baselines, we measured their ability to evaluate episodes independently and answer complex questions, without deliberately guiding them through the details of the story. 
% \subsection{Implementation Settings}
% Table~\ref{metrics} compares four different evaluation approaches for computing BLEU-4 scores. The first approach uploads original article files to the GPT-4o playground without additional processing. The second approach calls the GPT-4o application programming interface (API) and integrates the user’s framework to process the files before uploading. The third approach applies the same logic but uses GPT-4o-mini, and the fourth approach also calls the GPT-4o-mini API with the user’s framework. This layout clarifies how different usage modes may affect the final BLEU-4 outcome.
% \begin{table*}[htbp]
%     \centering
%     \caption{\label{metrics}
% Comparisons of four experimental settings for BLEU-4, contrasting direct file uploads with API-based approaches under both GPT-4o and GPT-4o-mini.
% }
%     \renewcommand{\arraystretch}{1.2}
%     \small
%     \begin{tabularx}{\textwidth}{lX}
%         \toprule
%         \textbf{Evaluation Method} & \textbf{BLEU-4} \\
%         \midrule
%         \rowcolor[HTML]{D0D0D0}GPT4o with framework 
%             & Upload Original Article files directly to GPT-4o playground \\
%         GPT4o playground
%             & Call the GPT-4o API, upload our processed files, and use our framework \\
%         \rowcolor[HTML]{D0D0D0}GPT4o-mini with framework
%             & Upload files directly to GPT-4o-mini playground \\
%         GPT4o-mini playground
%             & Call the GPT4o-mini API, upload our processed files, and use our framework \\
%         \bottomrule
%     \end{tabularx}
% \end{table*}

\subsection{Main Results}
As shown in Table~\ref{tab:performance}, our experiments demonstrated that the proposed framework significantly improved the detection of narrative inconsistencies. Evaluations using the framework were able to more accurately detect inconsistencies in character actions or plot progression. The retrieval-augmented generation process helped GPT better filter irrelevant information, understand the current story context, and improved its ability to detect narrative continuity across multiple episodes. When quantitatively compared to baseline methods, such as using GPT model alone, the proposed framework showed substantial improvements in evaluation accuracy.

\subsection{Ablation Studies}
\textbf{LLM type.} Table~\ref{tab:performance} shows that SCORE improves performance across all model families, with especially large gains for open-source models. Consistency increases up to 7.8\% (Llama-13B), coherence up to 7.7\% (GPT-4o), and item status recognition approaches 98\%. Open-source models (e.g., Qwen-14B, Llama-13B) see the largest QA gains (30–38 points), while commercial models (e.g., GPT-4, Claude 3) also benefit, though more modestly. Notably, SCORE boosts GPT-4o beyond GPT-4 in coherence, underscoring its scalability.  

\noindent\textbf{Configuration analysis.} Table~\ref{tab:ablation} shows that Dynamic Tracking is critical, with its removal causing the largest drops in consistency (–17.6 points) and item accuracy (–37.1 points). Context Summary is similarly vital for coherence (–22.5\%). Hybrid Retrieval and Sentiment contribute positively but with smaller impacts.  

\subsection{Case Study}
We generated several dozen stories with SCORE, averaging 555 tokens and 13.2s runtime per episode. Evaluation across four metrics—character consistency, plot progression, emotional realism, and continuity—yielded average scores of 3–4/5 (Figure~\ref{fig:scores_distribution}), indicating generally cohesive narratives.

\section{Conclusion}
We present \textsc{SCORE}, a novel LLM-based framework that enhances long-term coherence and emotional consistency in AI narratives. By combining Dynamic State Tracking, Context-Aware Summarization, and Hybrid Retrieval within a RAG pipeline, SCORE achieves substantial improvements in coherence, stability, and hallucination reduction across multi-genre datasets. Its modular design ensures scalability and multi-LLM compatibility, though challenges remain in retrieval accuracy and computational efficiency.

\vfill\pagebreak
\bibliographystyle{IEEEbib}
\bibliography{strings,refs}

\end{document}